\documentclass{egpubl}
\usepackage{eg2024}

 \Arxiv   
\CGFStandardLicense

\usepackage[T1]{fontenc}
\usepackage{dfadobe}  
\usepackage{amsmath} 
\usepackage{pifont}
\usepackage{cite}  

\usepackage{cite}  
\BibtexOrBiblatex
\electronicVersion
\PrintedOrElectronic
\ifpdf \usepackage[pdftex]{graphicx} \pdfcompresslevel=9
\else \usepackage[dvips]{graphicx} \fi

\usepackage{egweblnk} 
\usepackage{lineno}
\usepackage{caption}
\usepackage{subcaption}
\usepackage{amssymb}

\title[Stylized Face Sketch Extraction via Generative Prior with Limited Data]%
      {Stylized Face Sketch Extraction via \\Generative Prior with Limited Data}
\author[Yun et al.]
{\parbox{\textwidth}{\centering Kwan Yun$^{*1}$\orcid{0000-0002-0205-5203} \quad\quad   Kwanggyoon Seo$^{*1}$\orcid{0000-0003-0570-4915}  \quad\quad 
         Chang Wook Seo$^{*1}$\orcid{0000-0002-3809-9515}  \quad\quad  
         Soyeon Yoon$^{1}$\orcid{0009-0008-9805-9145}\\
         \vspace{0.15cm}
         Seongcheol Kim$^{1}$\orcid{0009-0003-4277-854X} \quad\quad  
         Soohyun Ji$^{2}$\orcid{0009-0002-7506-3316} \quad\quad  
         Amirsaman Ashtari$^{1}$\orcid{0000-0001-5352-7306} \quad\quad  
         Junyong Noh$^{1}$\orcid{0000-0003-1925-3326}\\
         \vspace{0.3cm}
    {\parbox{\textwidth}{\centering 
        $^1$KAIST, Visual Media Lab \quad $^2$Technical University of Munich\\
    }
}
}}

\begin{document}

\captionsetup{labelfont=bf,textfont=it}

\newcommand{\fixq}[1]{\textcolor{black}{#1}}
\newcommand{\fixw}[1]{\textcolor{black}{#1}}
\newcommand{\fixe}[1]{\textcolor{black}{#1}}
\newcommand{\fixr}[1]{\textcolor{black}{#1}}

\newcommand{\news}[1]{\textcolor{black}{#1}}
\newcommand{\newa}[1]{\textcolor{black}{#1}}

\newcommand{\newq}[1]{\textcolor{black}{#1}}
\newcommand{\neww}[1]{\textcolor{black}{#1}}
\newcommand{\newe}[1]{\textcolor{black}{#1}}
\newcommand{\newr}[1]{\textcolor{black}{#1}}

\newcommand{\kg}[1]{\textcolor{black}{[\textbf{kg}]:~#1}}

\teaser{
 \vspace{-0.8cm}
 \includegraphics[width=0.8\linewidth]{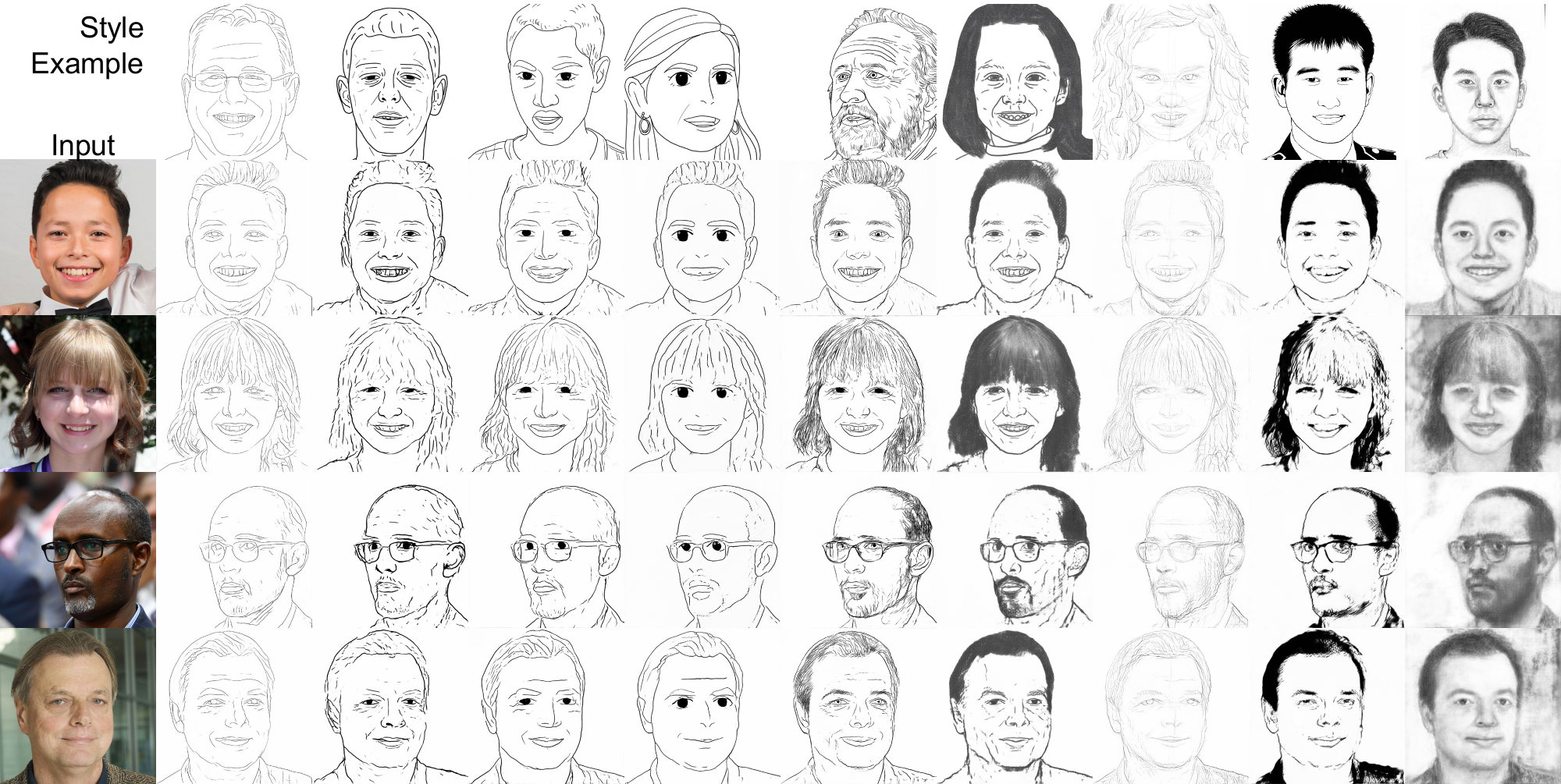}
 \centering
  \caption{Sketches extracted in nine different styles from the input images. Each style is trained with 16 pairs of data.}
\label{fig:teaser}
}

\maketitle

\begin{abstract}
Facial sketches are both a concise way of showing the identity of a person and a means to express artistic intention. While a few techniques have recently emerged that allow sketches to be extracted in different styles, they typically rely on a large amount of data that is difficult to obtain. Here, we propose StyleSketch, a method for extracting high-resolution stylized sketches from a face image. Using the rich semantics of the deep features from a pretrained StyleGAN, we are able to train a sketch generator with 16 pairs of face and the corresponding sketch images. The sketch generator utilizes part-based losses with two-stage learning for fast convergence during training for high-quality sketch extraction. Through a set of  comparisons, we show that StyleSketch outperforms existing state-of-the-art sketch extraction methods and few-shot image adaptation methods for the task of extracting high-resolution abstract face sketches. We further demonstrate the versatility of StyleSketch by extending its use to other domains and explore the possibility of semantic editing. The project page can be found in \href{https://kwanyun.github.io/stylesketch_project}{https://kwanyun.github.io/stylesketch\_project}.

\ccsdesc[500]{Computing methodologies~Artificial intelligence}
\ccsdesc[500]{Computing methodologies~Computer vision}
\ccsdesc[500]{Computing methodologies~Computer vision}

\printccsdesc

\end{abstract}
\vspace{-2mm}

\def\thefootnote{*}\footnotetext{These authors contributed equally to this work}\def\thefootnote{\arabic{footnote}}


\section{Introduction}

Sketching is the first step in artistic creation. It is a fundamental task for abstracting and expressing artistic ideas, which pre-visualizes the main structure and content of the final image. Despite the simple construction based on lines, sketches can have unique styles; some artists prefer to use lines with varied thicknesses depending on the locations of the constituent lines while others like to emphasize specific areas such as eyes with filled-in black color.

\fixr{Facial sketches can play a more important role than sketches that depict general objects. Facial sketch can show the identity of a person in a concise way while expressing artistic intention.} One recent line of research has attempted to reconstruct realistic human face images from sketches for various applications such as crime investigation, character design, educational training, and so on~\cite{chen2020deepfacedrawing}. In many of these applications, performance has been noticeably improved by using \fixe{generative models such as generative adversarial networks (GANs) and diffusion probabilistic models}~\cite{reffacenet,linestofacephoto,lee2020reference,chen2020deepfacedrawing,seo2022midms,portenier2018faceshop,liu2022deepfacevideoediting}.
However, these learning-based methods require a large \fixe{number} of paired sketch and photo images for training to produce adequate results.

\begin{figure}[htb]
  \centering
  \includegraphics[width=0.95\linewidth]{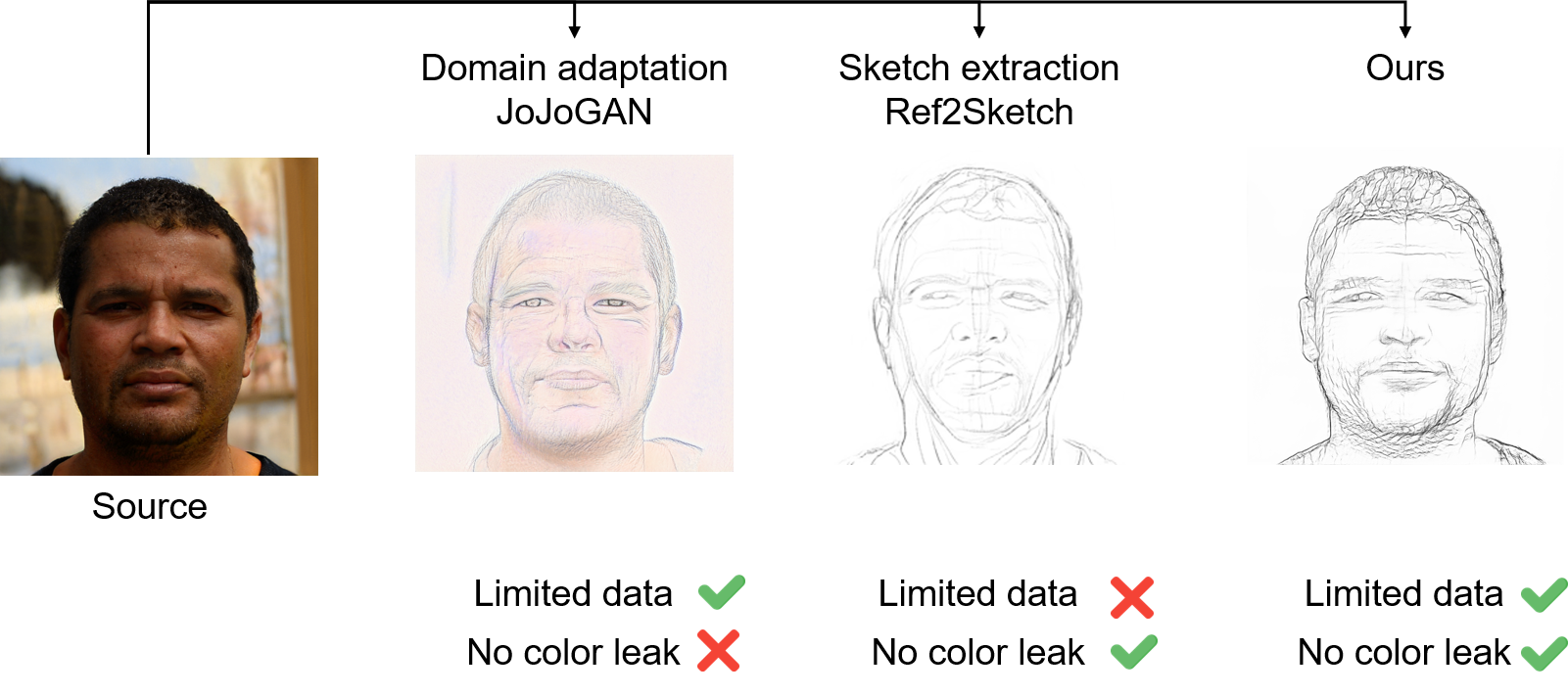}
  %

  \caption{Data-Artifact Tradeoff: Domain adaptation methods may require less data but often exhibit color bleeding~\cite{chong2022jojogan}. Image translation-based sketch extraction methods avoid such artifacts, but require a large dataset \cite{ashtari2022reference}. Our approach, using deep features from generative models while training from scratch, avoids artifacts and can be trained with a limited number of data.}

   \label{fig:difference}
\end{figure}

Instead of drawing sketches from scratch to obtain the dataset, many researchers have investigated how to extract synthetic sketches from a given image. Automatic techniques such as filter-based edge detection methods~\cite{canny1986computational,winnemoller2011xdog} or data-driven sketch extraction~\cite{lllyasviel2017sketchKeras,a2s,simo2016learning,ashtari2022reference,chan2022learning} methods have been proposed. Each type of sketch extraction method has its own unique characteristics. Filter-based edge detection methods are efficient, yet the results often consist of noisy and dotted lines. This is because their purpose is to find the precise edges instead of expressing sketch-like drawings. While learning-based sketch extraction methods can generate artistic style sketches based on training data, \fixe{they again} require a large dataset to train the system.

To address this challenge, we propose StyleSketch, which can extract high-resolution artistic sketches from a given face image after being trained with a limited number of data. StyleSketch leverages the deep generated features from the pretrained StyleGAN~\cite{karras2019stylegan2} as a building block. The deep features have been shown to have meaningful information for downstream tasks~\cite{zhang2021datasetgan, li2022bigdatasetgan,peebles2022gan,abdal2021labels4free}. Using these deep features as input, we trained a new sketch generator with deep fusion modules.

\news{StyleSketch is designed specifically for a sketch extraction task. A dedicated sketch generator is trained on the features generated from the pretrained StyleGAN. This makes StyleSketch different from existing domain adaptation methods.} Domain adaptation methods either fine-tune~\cite{chong2022jojogan,xiao2022few,zhu2021mind,ojha2021few} or employ adaptation modules to manipulate the weights of the pretrained generator~\cite{kim2022dynagan,yang2022pastiche}. While these approaches can be applied to many style related tasks, their stylized sketch extraction results are often unsatisfactory due to mode collapse or color bleeding. StyleSketch avoids these issues by designing a dedicated sketch generator that takes deep generated features of a pretrained generator as input. In addition, thanks to the rich semantic features of deep generated features, StyleSketch does not require a large training dataset, unlike other sketch extraction methods~\cite{yi2019apdrawinggan, yi2020line, yi2020unpaired, chan2022learning, ashtari2022reference, seo2023semi}. The differences of our method are highlighted in Fig.~\ref{fig:difference}

In addition to the newly proposed model architecture, we designed a specialized discriminator and \fixe{loss functions} for each part of the semantic region because the sketch style may differ from one semantic part to another. We also included a scheduler to guide the training process for faster convergence of the model. 
After training the sketch generator, StyleSketch can perform sketch extraction by inverting the image into the latent code of the StyleGAN~\cite{abdal2020image2stylegan++,richardson2021encoding,tov2021designing}. The latent code is then transformed to deep features that are fed to the sketch generator to produce an artistic sketch image. In addition to sketch extraction, StyleSketch can perform related tasks such as photorealistic image generation from a face sketch and semantic editing of sketch images.

For validation, we construct a high-resolution dataset of paired face sketches, \textbf{SKSF-A} featuring the same identity in seven different styles drawn by artists. There are several face image datasets that include the corresponding sketch drawings; however previous face sketch datasets consist of sketches with only one or at most a few drawing styles and the size of the dataset is typically small. Our newly constructed SKSF-A dataset consists of 938 images with seven different drawing styles and contains attributes for photo information. Our novel dataset is of higher resolution (1024$\times$1024) compared to existing face datasets which are typically lower than 779.62 $\pm$ 15.05 $\times$ 812.10 $\pm$ 13.92 in resolution~\cite{yi2019apdrawinggan,cufsf,cufs}. With this SKSF-A and previous face sketch datasets~\cite{yi2019apdrawinggan,cufs}, we show that our method can be trained with a limited number of paired data and performs equally or even better compared to state-of-the-art methods trained with more data. Example results are presented in Fig.~\ref{fig:teaser}.

Our contributions can be summarized as follows.
\begin{itemize} 

\item \news{We propose \textbf{StyleSketch}, a novel method that utilizes the pretrained StyleGAN features to extract sketches from face images enabling the network to be trained with a limited number of data.}

\item \news{StyleSketch utilizes a dedicated sketch generator that allows to aggregate the generated features effectively using a deep fusion module. During the training of the sketch generator, we use a two-stage learning process for improved convergence.}

\item In addition to sketch extraction, StyleSketch can perform the task of semantics editing of input sketches having diverse styles.

\item \news{We introduce \textbf{SKSF-A}, a high-resolution dataset of paired
face sketches in seven distinct styles drawn by professional artists. SKSF-A contains 134 identities and 938 sketches in total. SKSF-A is the most diverse and highest-resolution sketch dataset. }

\end{itemize} 

\section{Related work}
\subsection{Sketch Extraction}

The simplistic form of sketch extraction is edge detection. Most early studies of edge detection are based on techniques of finding the edge from the sudden change of color or brightness \cite{canny1986computational,winnemoller2011xdog}. While these methods are simple and powerful without learning from a large corpus of data, the generated results sometimes consist of noisy dots and scattered lines. To reproduce natural expression of drawings in the extracted sketches, learning-based approaches have been proposed. These methods are capable of detecting object boundaries or extracting lines in particular styles~\cite{a2s,xie2015holistically,lllyasviel2017sketchKeras,li2019photo,li-2017-deep}. 
Learning To Generate Line Drawings (Learn-to-draw) \cite{chan2022learning} improved previous methods by utilizing the depth and semantic meanings of a color image to extract a higher quality sketch. More recently, Ref2sketch \cite{ashtari2022reference} was introduced which allows a sketch to be extracted in a style of the reference sketch using paired training while Semi-Ref2sketch \cite{seo2023semi} utilizes contrastive learning for semi-supervised training. 

Unlike general purpose sketch extraction methods, some methods aim to extract sketches for a certain domain of images.
Due to the need for face sketches in many applications, there have been approaches specifically designed for facial sketch generation~\cite{yi2019apdrawinggan,yi2020unpaired,yi2020line}. 
These methods can generate higher quality sketch results compared to general sketch extraction methods due to the use of domain-specific networks and losses. 

Because of the nature of learning-based image-to-image translation methods, the aforementioned techniques require a large number of data to train the model.
Obtaining authentic sketch drawings of various styles with a large number of paired color images is very difficult and expensive, leading to data scarcity and consequent difficulty in training a sketch extraction model. To address this challenge, our method is designed to train the network with a few paired data by utilizing a deep generative prior in the face domain.

\begin{figure*}[!htb]
  \centering
  \includegraphics[width=0.95\linewidth]{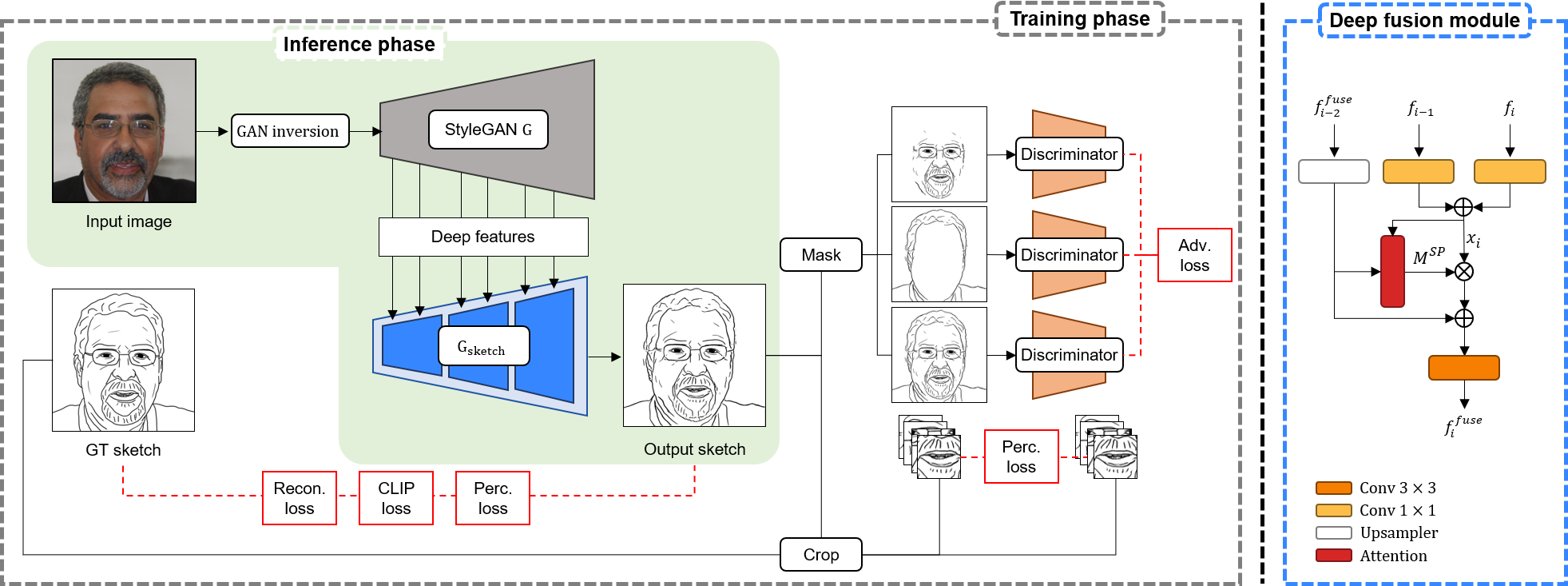}
  %

  \caption{\newe{Overview of StyleSketch. To extract a sketch from a face image, we first train a sketch generator $G_{sketch}$ that accepts deep features as input. $G_{sketch}$ is constructed with the deep fusion module represented on the right side of the figure. After training, we extract sketches by performing GAN inversion of the input image followed by feeding the deep features into $G_{sketch}$.}
  }
  \vspace{-0.5cm}
   \label{fig:overview}
\end{figure*}

\subsection{Few-shot Domain Adaptation}
While many attempts have been made to train generative models from scratch using a large corpus of data~\cite{karras2019stylegan,brock2018large,ho2020denoising,rombach2022high,chang2023muse}, several methods have shown promising results by leveraging a pretrained deep generative model to perform domain transfer with a limited number of data. 
The current approaches typically involve training a base model on a large number of source domain data and then fine-tuning it in a closely related target domain. One line of the approach utilizes additional modules in the generator to produce a target domain image~\cite{noguchi2019image,wang2020minegan,yang2022pastiche,sohn2023styledrop,zhang2023adding}.
Another approach employs regularization to avoid mode collapse that is often observed in many few-shot training approaches~\cite{ojha2021few,xiao2022few,mo2020freeze,li2020few}. More recently, there have been attempts to use the semantic power of a vision-language model CLIP~\cite{radford2021learning} that can guide the training process to transfer the source to the target domain while preserving the source distribution~\cite{gal2022stylegan,zhu2021mind}. \news{Optimizing the strokes with CLIP in vector representation has also been proposed~\cite{vinker2022clipasso} }. Instead of fine-tuning the generator, DynaGAN~\cite{kim2022dynagan} trains a hypernetwork using contrastive learning to modulate the weight of the generator for training with a limited number of stylized data while preventing mode collapse. 

These domain adaptation methods generally work well when the joint probability distribution of the target data is the same as or similar to the original source distribution \cite{han2020deep,zhang2020discriminative,courty2017joint,farahani2021brief}.
However, when the target domain is far from the original domain, it is challenging to learn and generate the target, such as sparse line drawings of a face where most regions have no information while real faces are pre-trained in the original domain. 
Recently, there have been attempts to personalize diffusion-based models by fine-tuning the denoising network~\cite{ruiz2023dreambooth} or finding a textual embedding given a concept or style image~\cite{gal2023textualinversion}.
We designed a model and a training scheme that can mimic the perceptual style of the artistic sketch images while preserving the geometric details and identity of the original photo.

\subsection{Leveraging StyleGAN Prior for the Downstream Task}
In addition to generating images in an unsupervised manner, StyleGAN~\cite{karras2019stylegan2} can be further extended to various downstream tasks.
pSp~\cite{richardson2021encoding} is a generic image-to-image translation framework that employs a frozen StyleGAN. The method inverts input images to the latent space of pretrained StyleGAN, which enables the generation of corresponding photorealistic images. In addition, GAN inversion methods~\cite{abdal2020image2stylegan++,richardson2021encoding,tov2021designing} can perform tasks such as image/video editing~\cite{shen2020interpreting, abdal2021styleflow, yao2021latent, tzaban2022stitch, xu2022videoeditgan, seo2022styleportraitvideo} and video generation~\cite{tian2021mocoganhd, fox2021stylevideogan}.

Another line of research has explored the \fixe{deep geometric features generated by pretrained GAN} models to exploit rich semantic information. 
DatasetGAN~\cite{zhang2021datasetgan} uses a few paired labeled data to train a semantic segmentation model. Labels4free \cite{abdal2021labels4free} uses a background and foreground generator to train an additional segmentation network in an unsupervised manner, and GANgealing \cite{peebles2022gan} utilizes GAN supervised training for dense visual alignment. These methods are possible because the features generated by StyleGAN hold valuable information that is structurally aligned with the task. Building upon this concept, we leverage the features generated from StyleGAN and put emphasis on the connectivity and style of the sketch edge.

\section{Methods}

\subsection{Overview}

StyleSketch extracts sketches from a given input face image through the process described below.

\begin{enumerate}  

\item First, a paired sketch and image dataset are prepared and the RGB image is projected into the pretrained latent space of StyleGAN $G$. With the projected latent code $w^+$, we hijack the deep features $F$ from StyleGAN. $F$ are extracted at the end of each style block resulting in 18 features total in different sizes.

\item The deep features and their corresponding sketch $I_{sketch}$ are used to train the sketch generator $G_{sketch}$. $G_{sketch}$ is trained in a supervised manner with additional part-based loss functions using a segmentation network. After training $G_{sketch}$, we can simultaneously generate both photo-realistic images from the original pretrained StyleGAN and sketches from $G_{sketch}$ of the same identity and structure.

\item To extract a sketch from an image at inference, we project the image into the latent space of pretrained StyleGAN and use the projected latent code to synthesize the sketch using $G$ and $G_{sketch}$.

\end{enumerate}
In the following subsections, we will describe the data construction (Sec.~\ref{subsec:dataprep}), the structure of sketch generator $G_{sketch}$ (Sec.~\ref{subsec:network}), its loss functions (Sec.~\ref{subsec:loss}), and finally the proposed sketch extraction method (Sec.~\ref{subsec:infer}).

\subsection{Data Construction}\label{subsec:dataprep}
Our goal is to train $G_{sketch}$ using a limited number of paired data by leveraging the semantics of a deep generative prior which is known to embed the knowledge of the human face. To use such prior, we first need to project the input image to the latent space of StyleGAN using a pretrained GAN inversion encoder~\cite{tov2021designing} and a few optimization steps~\cite{abdal2020image2stylegan++}. This latent code $w^+$ is then passed through StyleGAN, and we hijack the intermediate generated deep features as $F=\{f_1, ... f_n\}$ where $n$ is a total number of features.
We use the extracted deep features $F$ and their corresponding sketch $I_{sketch}$ as paired data to train the \fixe{sketch generator} $G_{sketch}$.

\subsection{Sketch Generator}\label{subsec:network}

Our sketch generator $G_{sketch}$ was built with convolution and attention blocks, as shown in Fig.~\ref{fig:overview}. $G_{sketch}$ takes $F$ as input and synthesizes the corresponding sketch $\hat{I}_{sketch}$, which can be written as follows:

\begin{equation}
\label{eq:sketchgen}
\begin{aligned}
& \hat{I}_{sketch} = G_{sketch}(F) . 
\end{aligned}
\end{equation}
Unlike other learning-based sketch extraction methods in which a network takes an image as input~\cite{ashtari2022reference,chan2022learning,yi2019apdrawinggan}, our method accepts multiple deep features that have different spatial resolutions and channel numbers. The width and height of the deep features increase gradually from $4\times4$ to $1024\times1024$ while the channel number decreases from $512$ to $3$. Because of the multiple inputs that have differences in spatial resolution and channel number, the deep features are gradually fused from coarse to fine levels.
Our deep fusion module accepts two features $(f_{i},f_{i-1})$ from the $(i,i-1)^{th}$ layer of pretrained StyleGAN, which have the same spatial resolution, along with previously fused features $f^{fuse}_{i-2}$.
The channel of $(f_{i},f_{i-1})$ is decreased using a $1\times1$ convolution layer for computational efficiency and the features are concatenated to obtain $x_i$. Because  $f^{fuse}_{i-2}$ has a lower spatial resolution compared to that of $(f_{i},f_{i-1})$, we first resize the features using an upsampling layer $\lceil \rceil$.

\news{
A spatial attention map $M^{sp}$ is synthesized from the fused feature representation $\lceil f^{fuse}_{i-2} \rceil$ and subsequently multiplied by $x_i$. $M^{sp}$ serves as the gating coefficient for input information $x_i$. The resulting output from this step is concatenated with $\lceil f^{fuse}_{i-2} \rceil$ and passed through a convolutional layer to produce $f^{fuse}_i$.} The deep fusion module can be written as follows:

\fixr{
\begin{equation}
\label{eq:1}
\begin{aligned}
x_i =& Conv^{1\times1}(f_{i}) \oplus Conv^{1\times1}(f_{i-1}) \\
f^{fuse}_{i} =& Conv^{3\times3} ( (M^{sp} \otimes x_i) \oplus \lceil f^{fuse}_{i-2} \rceil ),
\end{aligned}
\end{equation}
where $Conv^{3\times3}$ and $Conv^{1\times1}$ are the convolution layers,} $\otimes$ is element-wise multiplication, and $\oplus$ is concatenation in the channel dimension.

\subsection{Objective Functions}\label{subsec:loss}

To train the sketch generator $G_{sketch}$, we use four loss functions: reconstruction loss $\mathcal{L}_{1}$, perceptual loss $\mathcal{L}_{perc}$, clip loss $\mathcal{L}_{clip}$, and adversarial loss $\mathcal{L}_{adv}$.
When computing $\mathcal{L}_{perc}$ and $\mathcal{L}_{adv}$, we additionally use a segmentation model~\cite{yu2018bisenet} to mask out or crop parts of the face so that each semantic area can have a different style as performed in previous methods~\cite{yi2019apdrawinggan,chen2020deepfacedrawing}.

The most basic loss for an image-to-image translation task is to use a reconstruction loss. We use L1 distance between the ground truth sketch image $I_{sketch}$ and $\hat{I}_{sketch}$, which can be written as follows:
\begin{equation}
\label{eq:recon}
\begin{aligned}
\mathcal{L}_{recon} = \| I_{sketch} -\hat{I}_{sketch} \|_1.
\end{aligned}
\end{equation}
We use $\mathcal{L}_{recon}$ in the first stage of learning to make the network converge faster and perform better. \fixw{This will be further explained in the last paragraph of this section.}

The perceptual loss function $\mathcal{L}_{perc}$ guides the synthesized $\hat{I}_{sketch}$ to closely resemble the ground-truth $I_{sketch}$ using a pretrained VGG16~\cite{simonyan2014very}, which is shown to be effective in following the fine details of the target style~\cite{johnson2016perceptual,xu2019perceptual}. The loss is formulated as follows:
\begin{equation}
\label{eq:perceptual}
\begin{aligned}
\mathcal{L}_{perc} &= \sum_{l}\|\phi_{l}(I_{sketch}) - \phi_{l}(\hat{I}_{sketch})\|_1 \\
& + \frac{1}{N}\sum_{C}\sum_{l}\|\phi_{l}(C(I_{sketch})) - \phi_{l}(C(\hat{I}_{sketch}))\|_1,
\end{aligned}
\end{equation}
where $\phi_{l}$ refers to the activation map of the $l^{th}$ layer of a pretrained VGG16, and $C$ represents the cropping operation for each facial part such as eyes, nose, and lips, using a segmentation model~\cite{yu2018bisenet}. $N$ refer to the total number of cropped face features.

In addition to the perceptual loss, we use a visual-text embedding CLIP~\cite{radford2021learning} to capture high-level semantics for sketches, as reported in Learn-to-draw \cite{chan2022learning}. Thus, we compute the cosine similarity of the CLIP embedding between $I_{sketch}$ and $\hat{I}_{sketch}$ as follows:
\begin{equation}
\label{eq:clip}
\mathcal{L}_{clip} = 1 - \langle CLIP(I_{sketch}),CLIP(\hat{I}_{sketch}) \rangle, 
\end{equation}
where $\langle \cdot, \cdot \rangle$ is the consine distance function.

Lastly, we jointly train multiple discriminators with $G_{sketch}$ so that the generator accurately reflects the trained sketch style. By incorporating the discriminators, $G_{sketch}$ is enforced to generate a sketch image that belongs to the style observed in the ground truth sketch dataset. We use a conditional patch adversarial loss~\cite{isola2017image} as follows:

\begin{equation}
\label{eq:adv}
\begin{aligned}
\mathcal{L}_{adv} &=  \sum_{D} \Bigr[\mathbb{E}_{I,I_{sketch}}[ \log D(M_D \otimes I ,M_D\otimes I_{sketch})] \\
& +\mathbb{E}_{I,\hat{I}_{sketch}}[\log(1-D(M_D\otimes I,M_D\otimes \hat{I}_{sketch})]\Bigr], 
\end{aligned}
\end{equation}

where $M^{D}$ is a binary mask that indicates where the discriminator $D$ should focus generated from ~segmentation\cite{yu2018bisenet}.
We further jointly augment $I$, $I_{sketch}$, and $\hat{I}_{sketch}$ using the non-leaking augmentation method proposed in StyleGAN-ada \cite{karras2020styleganada}.

The final objective function can be written as follows:
\begin{equation}
\label{eq:total}
\begin{aligned}
G^*_{sketch} = \arg \min_{G_{sketch}} & \max_{D} \lambda_{recon} \mathcal{L}_{recon} + \lambda_{perc} \mathcal{L}_{perc} \\
& + \lambda_{clip} \mathcal{L}_{clip} +  \lambda_{adv} \mathcal{L}_{adv}, 
\end{aligned}
\end{equation}
where $\lambda_{recon}$, $\lambda_{perc}$, $\lambda_{clip}$, and $\lambda_{adv}$ are the weight for each loss term.

\subsection{Two-stage Learning}

\begin{figure}[!t]
  \centering
  \includegraphics[width=1.0\linewidth]{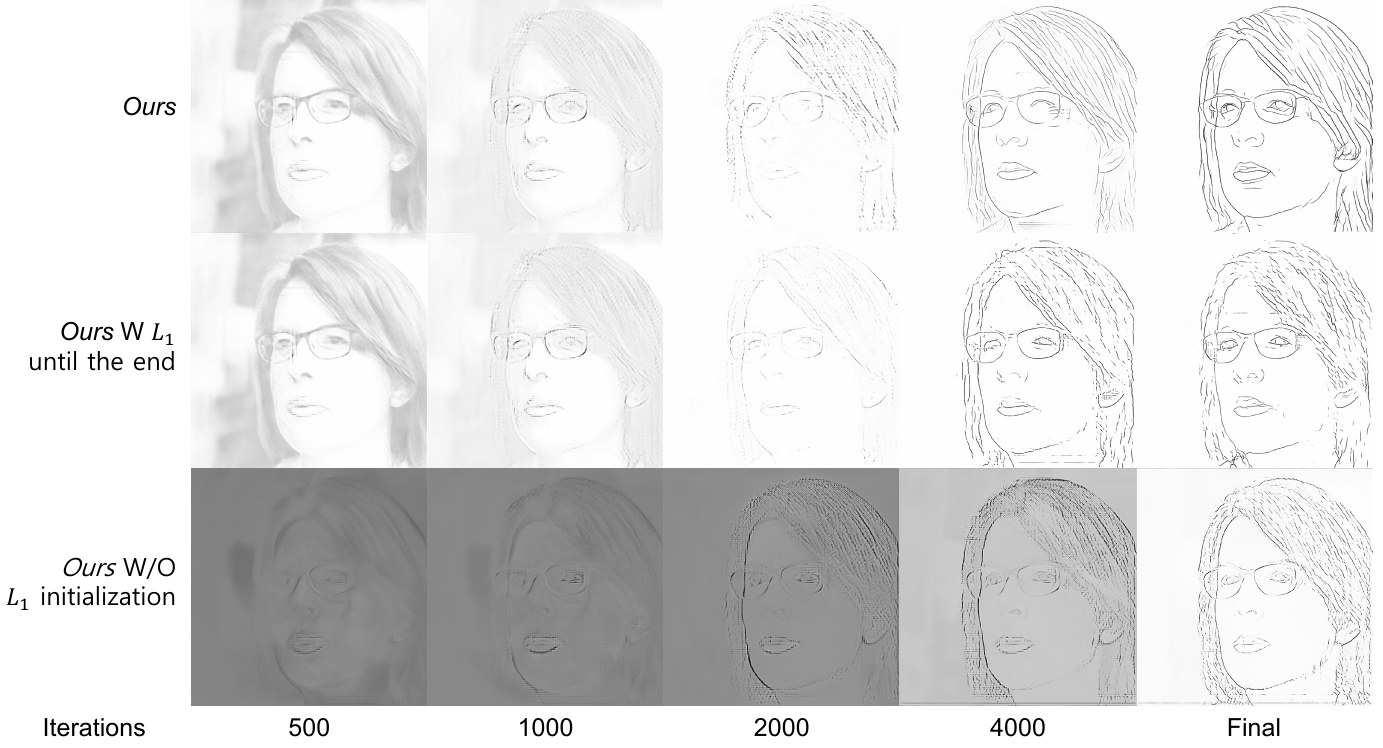}
  %

  \caption{\fixw{Results produced according to the iterations performed} with and without the initial stage. Training without L1 loss requires a large number of iterations to optimize the output into the sketch domain.}
   \label{fig:curriculum}
\end{figure}

Despite its wide use as a loss function, the L1 distance, which is $L_{recon}$ in our method, can be problematic for sketch extraction because it focuses more on pixel-wise differences than perceptual information~\cite{ashtari2022reference,xu2019perceptual}.
While $L_{recon}$ enables capturing of the overall structure of the sketch faster in the initial training steps, it can degrade the final quality when it is used throughout training, leading to noisy dotted patterns. 
To this end, we adopt a two-stage training approach that utilizes $L_{recon}$ in the initial stage only. As shown in Fig.~\ref{fig:curriculum}, with two-stage learning, our model converges faster in the early stage of training and ultimately leads to higher quality sketches. More experiments and their results can be found in Sec.~\ref{subsec:ablation}.

\subsection{Inference}
\label{subsec:infer}

Once the sketch generator $G_{sketch}$ has been trained, we can extract a sketch from a given face image by performing GAN inversion, as explained in Sec.~\ref{subsec:dataprep}. Specifically, the input image is transformed to a latent code \news{$w^+$} \fixr{that will be fed to the pretrained StyleGAN $G$ for the generation of deep features $F$.
$G_{sketch}$ then takes $F$ as input} to generate the corresponding sketch $\hat{I}_{sketch}$. The process is shown in the green box of Fig.~\ref{fig:overview}.


\newcommand{\cmark}{\ding{51}}%
\newcommand{\xmark}{\ding{53}}%

\section{SKSF-A Dataset}
\news{There exist several face image datasets that include the corresponding sketch drawings. We briefly describe the details of these datasets based on the descriptions of the FS2K paper~\cite{fan2022facial}. See Table~\ref{tab:dataset} for the summarized details.}

\news{
Among widely used face sketch datasets of a single drawing style, CUFS~\cite{cufs} consists of 606 color photo and sketch pairs, emphasizing shading effects created by pencil drawings. CUFSF~\cite{cufsf}, an extension of CUFS, is designed for evaluating face sketch synthesis methods and offers 1,194 pairs with varied contrast and background colors. APDrawing~\cite{yi2019apdrawinggan} provides high-quality face drawings, featuring 140 pairs expanded to contain 910 additional synthetic sketches through augmentations. These datasets~\cite{cufs, cufsf, yi2019apdrawinggan} include paired authentic sketches, enabling easy evaluation of image-to-image translation models.}

\begin{table}[]
\centering
\renewcommand{\arraystretch}{1.2} 
\caption{\label{tab:dataset}\news{Comparison with existing authentic face sketch datasets.}}

\resizebox{\linewidth}{!}{
\begin{tabular}{|l|c|c|c|c|c|c|}
\hline
Dataset                        & Total               & Style & Resolution                      & Attributes & Paired & Public \\ \hline
\textbf{SKSF-A(ours)}                   & 938                 & \textbf{7}   & \textbf{1024 $\times$ 1024}            &  \cmark      &    \cmark  &\cmark   \\
CUFS~\cite{cufs}                        & 606                 & 1            & 200 $\times$ 250                 &  \xmark    &  \cmark      &  \cmark      \\


CUFSF~\cite{cufsf}                      & 1,194                 & 1            & 779.62  $\times$ 812.10  & \xmark    &  \cmark  &  \cmark       \\


DisneyPortrait~\cite{disney}            & 672                 & 1            &             -              &    \xmark  &    \xmark    &   \cmark     \\
APDrawing~\cite{yi2019apdrawinggan}     & 140                 & 1            & 512 $\times$ 512                   &  \xmark  &   \cmark     &    \cmark    \\
UPDG~\cite{yi2020unpaired}              & 952                 & 3            &          -                    & \xmark  &  \xmark      &    \xmark    \\
FS2K~\cite{fan2022facial}               & \textbf{2,104}     & 3            & 299.74  $\times$ 273.56  &  \cmark   & \cmark  &     \cmark   \\ \hline

\end{tabular}
}
\end{table}

\news{
The Disney portrait dataset~\cite{disney} includes 672 sketches of a single style and in four levels of abstraction. UPDG~\cite{yi2020unpaired} comprises 952 sketches in three styles. Different from the aforementioned datasets~\cite{cufs, cufsf, yi2019apdrawinggan}, Disney portrait dataset~\cite{disney} and UPDG do not provide corresponding photo images, making it challenging to compare the images generated from image-to-image translation models with their corresponding ground truths. The most recent and largest FS2K dataset~\cite{fan2022facial} features 2,104 pairs of photos and sketches in three different styles created by different artists. The sketches underwent slight changes for artistic expression. Attributes such as hair style and facial feature colors are also provided. Unfortunately, the provided images are of low resolution and only one style is provided for each identity, which restricts its usability. }

\begin{figure}[h]
  \centering
  \includegraphics[width=1\linewidth]{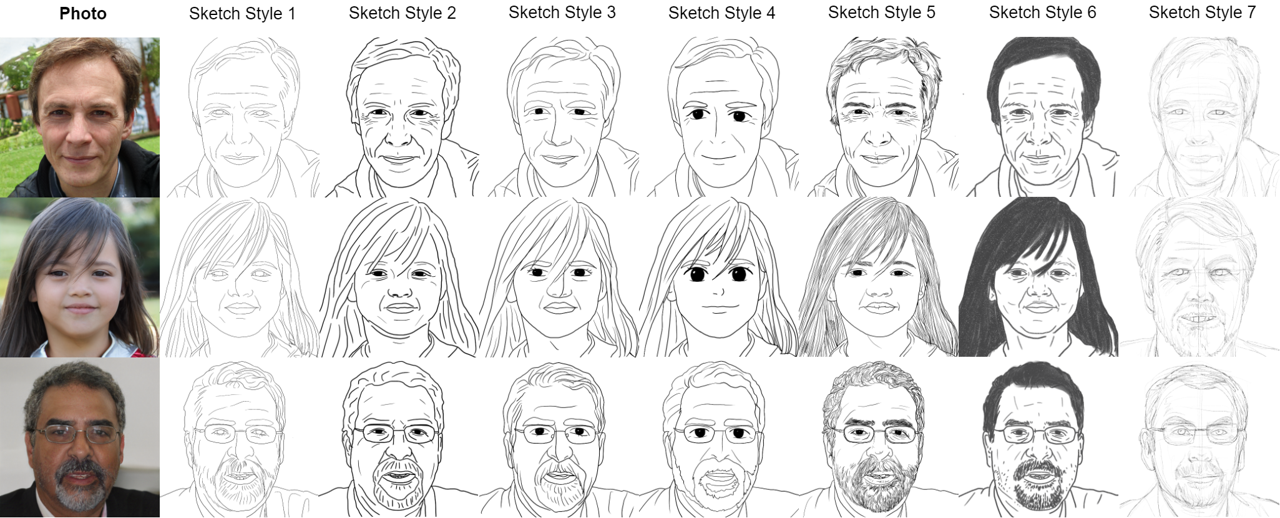}
  \caption{\label{fig:SKSF-A}
           Overview of our authentic face sketch dataset (SKSF-A). The dataset consists of photo and sketch pairs while the sketch has one of 7 different styles.}
    \vspace{-0.4cm}
\end{figure}

\news{Focusing on resolution, style variety, and a paired setting,  we build a dataset of authentic sketch style faces (SKSF-A) which consists of face sketches in seven different styles.
The seven different styles represent} complex and diverse details such as clear lines, rough lines, or filled-in colors. Because the SKSF-A dataset is drawn on FFHQ~\cite{karras2019stylegan} aligned face, it can be easily utilized with methods trained with the FFHQ dataset. See Fig.~\ref{fig:SKSF-A} for examples.

\subsection{Face Photos}

For free licensing of our SKSF-A dataset and its convenient distribution, we generated synthetic face photo images by utilizing GAN-based methods. Among various realistic face image generation methods \cite{sauer2022stylegan,humayun2022polarity,lin2021anycost,wang2022diffusion,karras2020analyzing,karras2020training}, we utilized StyleGAN2 with the weights pretrained by the FFHQ~\cite{karras2019stylegan} dataset to generate high quality face images. From the generated images, we selected 134 images without any artifacts on the face. To utilize these images as a learning-based method dataset, we divided the SKSF-A dataset into a training set of 80 images and a validation set of 54 images. In the experiments, our method uses 16 randomly selected images from the 80-image training set unless otherwise noted.

\subsection{Face Sketches}

Two artists participated in drawing the sketches for the SKSF-A dataset. These artists have no color-blindness or color-weakness. As shown in Fig.~\ref{fig:SKSF-A}, the sketches represent diverse styles.  Sketches of the sixth and seventh styles were drawn by the second artist and the others were drawn by the first artist. The artists drew all of the sketches with a pen tablet device and drawing software~\cite{clip_studio}.

\begin{figure*}[!htb]
  \centering
  \includegraphics[width=0.85\linewidth]{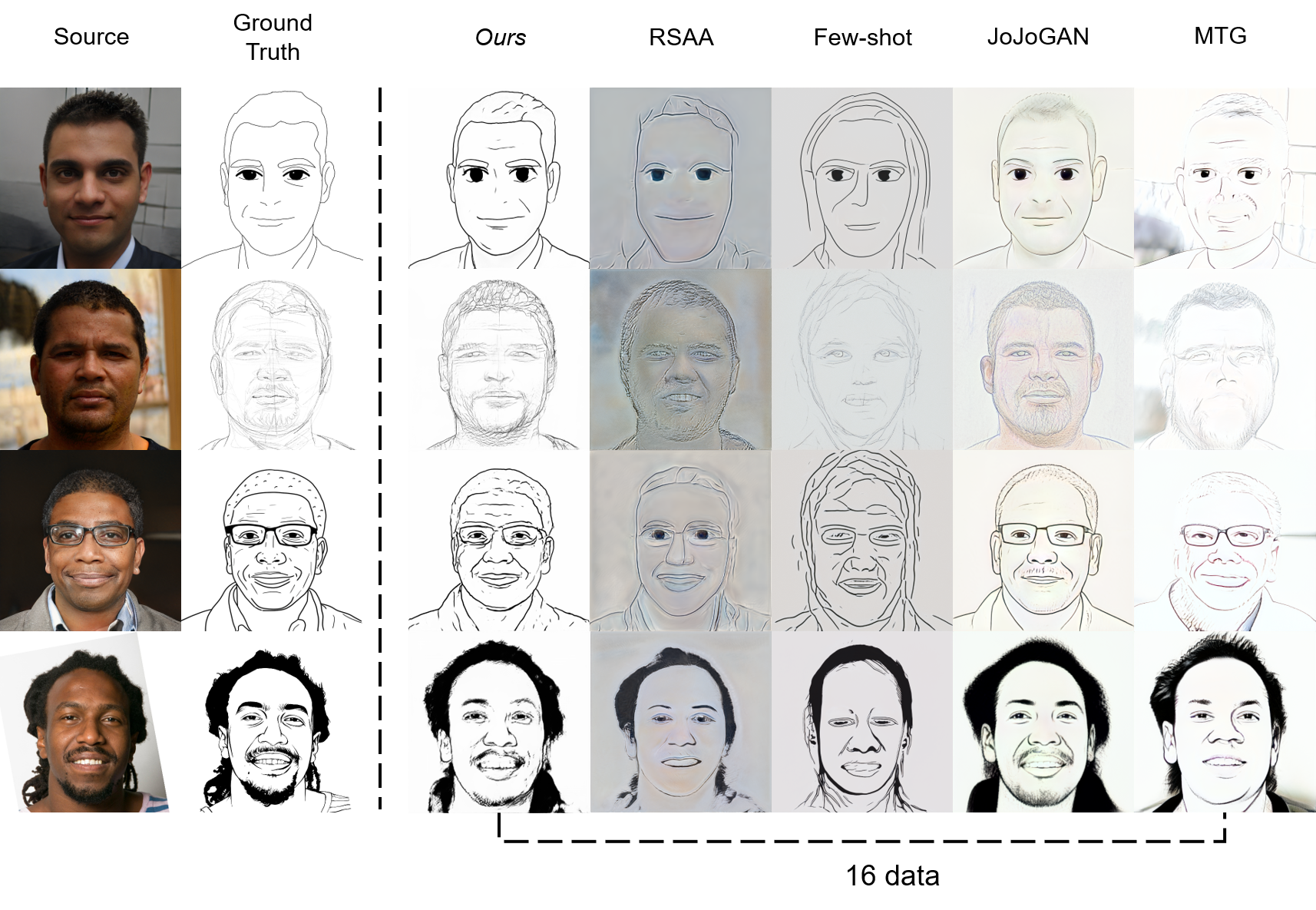}
  %
  \vspace{-0.5cm}
  \caption{\label{fig:comparison1}
           Comparison with domain adaptation baseline methods. All methods are trained with 16 data. Our method produced the best quality results among all the methods while other methods suffer from identity shift or color bleeding.}
    \vspace{-0.5cm}
           
\end{figure*}
\begin{figure*}[!htb]
    \centering
    \includegraphics[width=0.90\linewidth]{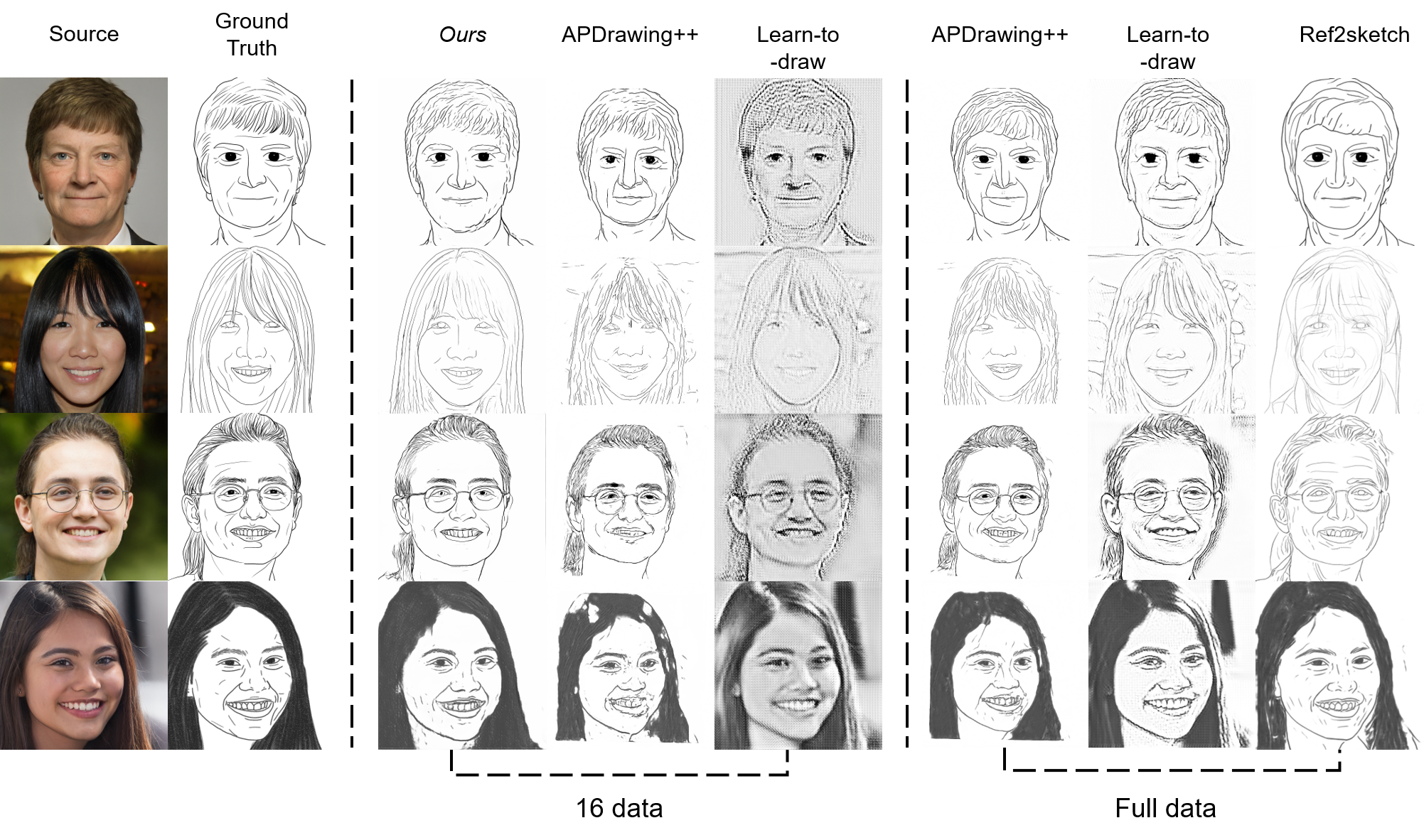}
    %
    \vspace{-0.5cm}
    \caption{\label{fig:comparison2}
           Comparison with sketch extraction baseline methods. Methods on the left are trained with a limited number of data (16) and on the right are trained with the full data (from 70 to 88 while 560 for Ref2sketch). Our method produced the best quality results among the methods trained with a limited number of data. The quality of our results is also comparable to that of the results produced by the models trained with the full data. }
    \vspace{-0.5cm}
           
\end{figure*}
\begin{figure*}[!htb]
    \centering
    \begin{subfigure}{0.49\linewidth}
         \centering
         \includegraphics[width=\linewidth]{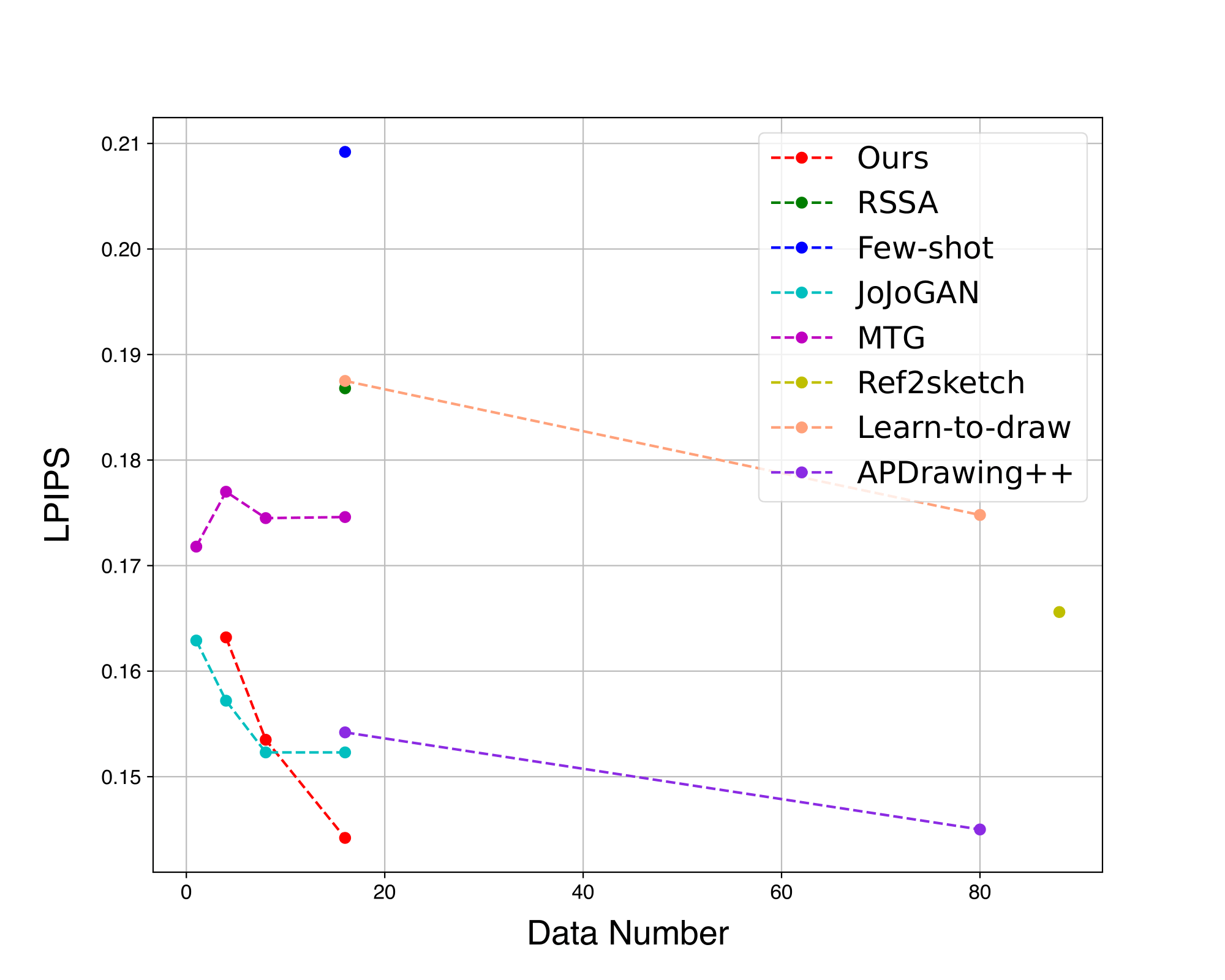}
         \label{fig:plot_lpips}
    \end{subfigure}
    \hspace{-10mm}
    \begin{subfigure}{0.49\linewidth}
         \centering
         \includegraphics[width=\linewidth]{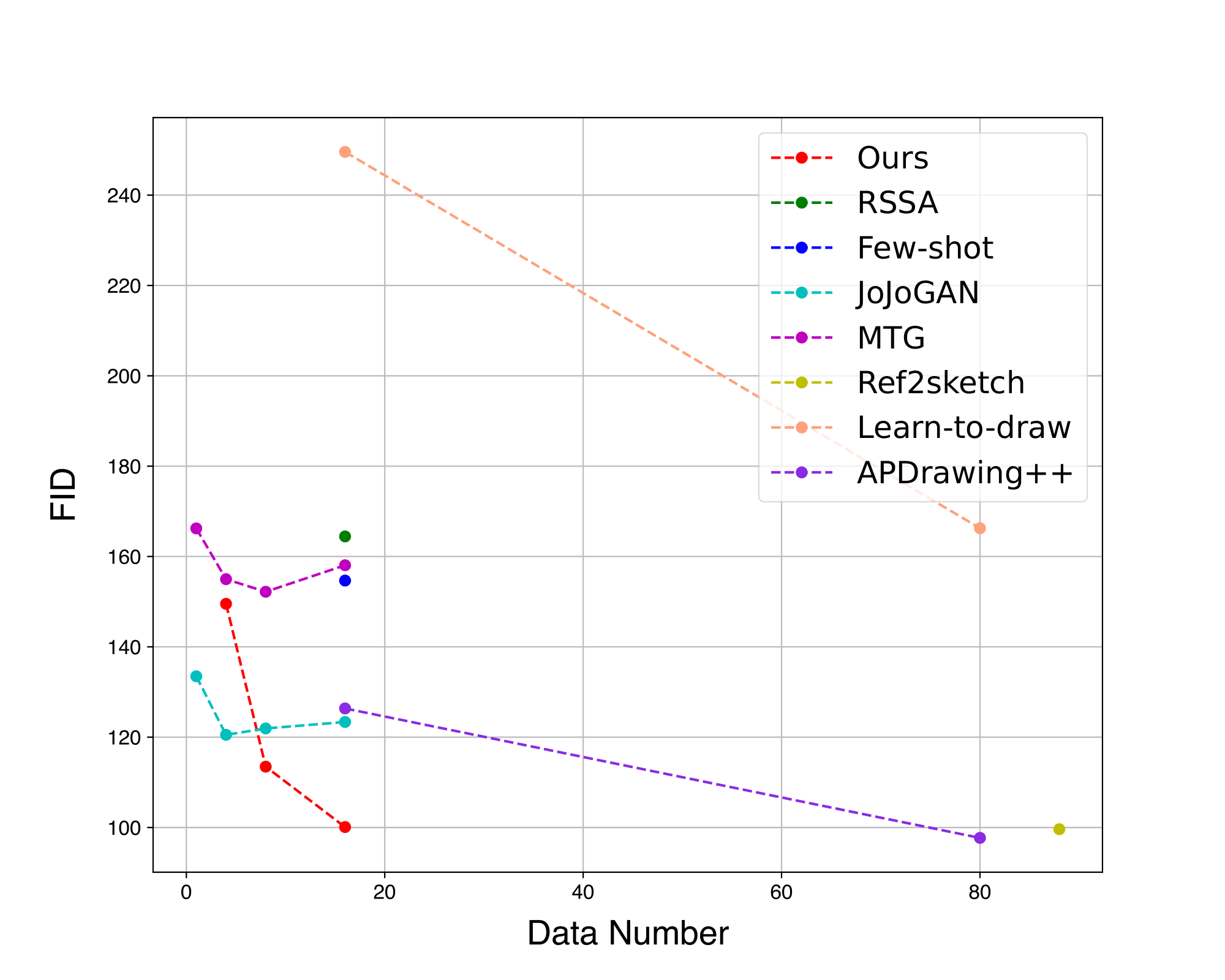}
         \label{fig:plot_lpips}
    \end{subfigure}
    \caption{\label{fig:plot} \news{Quantitative results. The left and right plots show the LPIPS and FID scores, respectively.}}
\end{figure*}

\section{Experiments}

\subsection{Implementation Details}
We implemented StyleSketch and trained the sketch generator on a computer with 2 Nvidia RTX-3090 GPUs. All experiments were conducted on a computer with an Nvidia RTX-3090 GPU.
In all of our experiments, StyleGAN2 pretrained on FFHQ~\cite{karras2020analyzing} was used unless noted otherwise. For training $G_{sketch}$, the learning rate was fixed to 0.00014. The parameters used in Eq.~\ref{eq:total} were fixed at $\lambda_{adv}$ = 1, $\lambda_{clip}$ = 120, $\lambda_{perc}$ = 1.2 for \fixe{both the initial stage and the second stage,} and $\lambda_{L1}$ = 200 for the initial stage. Out of a total of 7200 iterations, the first 1600 iterations were considered the initial stage. ~\news{The sketch generator requires a total of 9 hours for training and 0.43 seconds for inference.}
\newr{We used 16 pairs of face and sketch images for training unless noted otherwise.}

\subsection{Comparison with Baselines}
For quantitative evaluation, we compared our method with seven different baselines. Four baselines are domain adaptation methods for stylization while three methods are sketch extraction methods. Mind the gap (MTG)\cite{zhu2021mind} and JoJoGAN \cite{chong2022jojogan} are one shot domain adaptation methods; the former fine-tunes the StyleGAN \cite{karras2020analyzing} with bidirectional CLIP information and the latter achieves the same with augmentation. RSSA \cite{xiao2022few} and few-shot image generation (Few-shot) \cite{ojha2021few} are few shot adaptation methods that utilize a pretrained StyleGAN to manipulate face images. RSSA uses relaxed spatial structure alignment while Few-shot \cite{ojha2021few} utilizes a cross-domain distance consistency loss. These features help prevent overfitting when training with a limited number of data. All four domain adaptation methods were trained in an unpaired setting with proper alignment. However, this approach sometimes failed to align sparse sketch styles. Therefore, we aligned the sketches using our paired method for training.

Ref2sketch~\cite{ashtari2022reference} is a general purpose sketch extraction model which allows \fixw{line drawings of} various styles to be extracted in one model by imitating the styles of a reference input. Learning to generate line drawing (Learn-to-draw) \cite{chan2022learning} is also a general purpose sketch extraction model \fixw{that} utilizes depth and semantic meaning of images to enhance the quality of output. \fixw{These} methods are not specialized on generating facial sketches; therefore, we also chose APdrawing++ \cite{yi2020line} as \fixw{another} baseline. This method also utilizes a pretrained model related to face image features and is specially designed to generate face sketch images. 

For our analysis, we used three different face sketch datasets: SKSF-A, APDrawing~\cite{yi2019apdrawinggan}, and CUFS~\cite{cufs}. SKSF-A consists of seven different styles, and we trained the models separately for each style, except for Ref2sketch, which was designed to train various styles within a single model. To ensure a fair comparison with the state-of-the-art method APdrawing++, we aligned all testing images using five facial landmarks and transformed them to match the control points predefined by APdrawing++. Additionally, we trained Learn-to-draw and APdrawing++ using two different datasets: one with \fixw{a} limited \fixw{number of} data, similar to our method (16 images), and the other with the full dataset (70-88 images) as originally provided. Few-shot was trained \fixw{only} with a limited \fixw{number of data reflecting the specific characteristic of the model.} \fixw{Similarly,} Ref2sketch was trained with the full dataset. Other settings such as optimization and \fixw{the number of} iterations were \fixw{identically employed} according to the \fixw{descriptions in the} paper and code \fixw{of each method}. \newr{While \textit{Ours} and other domain adaptation methods can extract sketches in high resolution of 1024$\times$1024, other methods do so in lower resolutions. Specifically, APDrawing++ produces results in 512$\times$512, and Ref2Sketch and Learn-to-draw produce results in 256$\times$256.}

As evaluation metrics, we used LPIPS~\cite{zhang2018unreasonable} for perceptual similarity and FID~\cite{heusel2017gans} for distribution similarity. Because FID requires a large number of test sets, we used it as a secondary indicator, following the approach taken in previous sketch extraction research ~\cite{yi2019apdrawinggan,yi2020line,ashtari2022reference,seo2023semi}.
Figs.~\ref{fig:comparison1} and \ref{fig:comparison2} show the example results, and \newa{Fig.~\ref{fig:plot}} shows evaluation scores. Ours achieved the best scores when a limited number of data were used, outperforming the baseline methods including the ones trained with full data. We provide the tabulated complete results of all nine styles \newa{in Table 2 of the supplementary material.}
More examples from the methods \cite{yang2022pastiche,seo2023semi,gal2023textualinversion,ruiz2023dreambooth,zhang2023adding} that could not generate sketches in a trained style or produced severe artifacts are presented \newa{in Sec. 4.3 of the supplementary material.}

\subsection{Perceptual Study}
We performed a perceptual study to evaluate the output sketches based on human perception with different sketch extraction methods. \fixw{We recruited} 53 participants \fixw{for} a survey \fixw{that compared} generated sketches with test images from three different datasets: SKSF-A, CUFS, and APDrawing. \neww{Because we did not prioritize any particular groups of individuals, our assessments should not be influenced by the details of any demographic distribution.} \fixw{A total of 20 comparison tasks for randomly chosen results were presented to each participant. Among the 20 comparisons,} 16 comparison \fixw{data were} from the SKSF-A dataset \fixw{that} consists of 7 different styles, 3 comparison \fixw{data were} from the APDrawing dataset, and 1 comparison \fixw{data was} from the CUFS dataset. The participants were presented with a ground truth target sketch image along with four generated sketches for each comparison. \fixw{Note that all of} the baseline methods from sketch extraction methods ~\cite{ashtari2022reference,yi2020line,chan2022learning} were trained using the full dataset, whereas our method was trained using only a limited number of data.

\begin{figure}[t]
  \centering
  \includegraphics[width=1.00\linewidth]{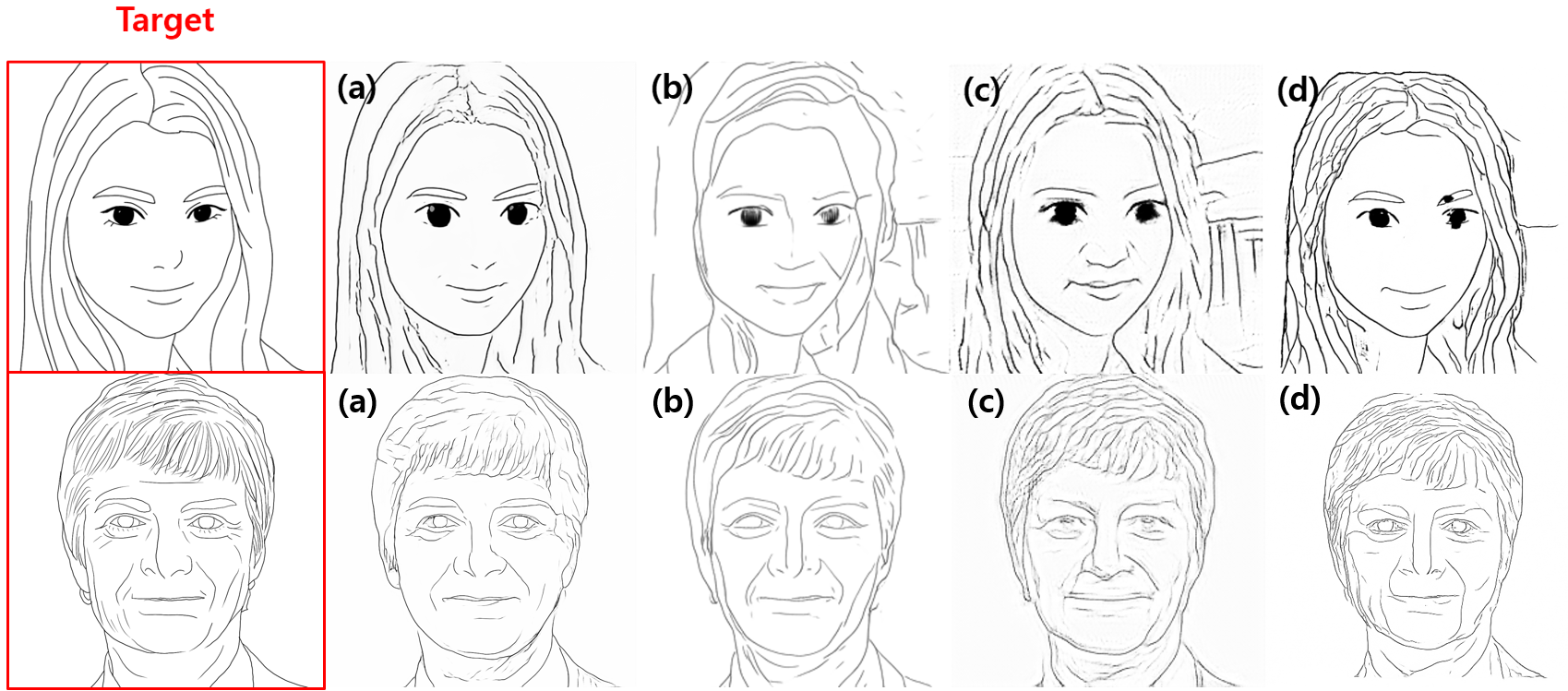}
  \caption{\label{fig:perceptual} Comparison sample of our perceptual study. The \fixw{sketch} results shown above were produced by our method and three baseline methods: (a)
\textit{Ours}, (b) Ref2sketch, (c) Learn-to-draw, and (d) APDrawing++. The order in which these results are displayed was randomly selected for each comparison.}
\vspace{-0.3cm}
\end{figure}

The participants were asked to choose the \fixw{sketch} that most closely resembled the target \fixw{sketch} image. There was no time limit imposed on the participants during this process. An example of survey is depicted in Fig.~\ref{fig:perceptual}, and the resulting scores are listed in Table~\ref{tab:perceptual}. The outcome of this perceptual study demonstrates that our method trained with a limited \fixw{number of} data \fixw{produced results that were dominantly selected by the participants as that most similar to the target sketch image despite the fact that the competing} baselines were trained with the full data.

\begin{table}[t]
\centering
\caption{Results from the user perceptual study. \fixw{The percentage indicates the selected frequency.}}
\label{tab:perceptual}
\begin{tabular}{|l|c|}\hline 
\multicolumn{1}{|c|}{Method} & \textbf{User Score}     \\ \hline
\textit{Ours}                          & \textbf{ 42.92\%} \\ 
Ref2sketch                          &  24.43\%\\ 
Learn-to-draw                    &  11.33\%\\ 
APDrawing++                           &  21.32\%\\ \hline 
\end{tabular}
\vspace{-0.5cm}
\end{table}

\begin{figure*}[!htb]
  \centering
  \includegraphics[width=1\linewidth]{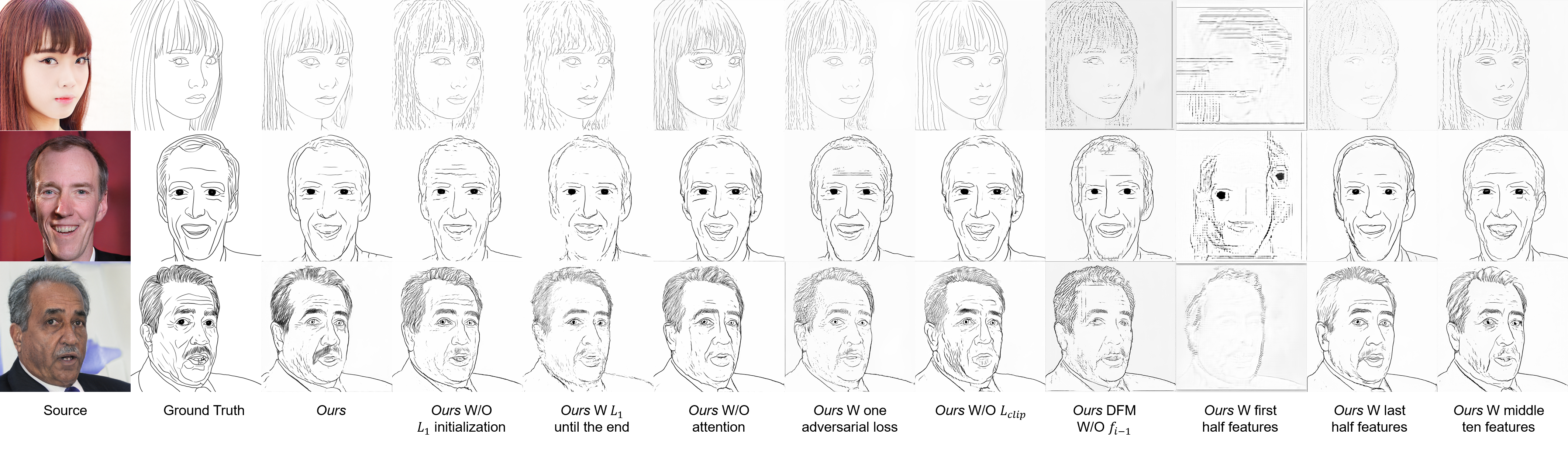}

  \caption{\label{fig:ablationexample}
           \fixw{Visual e}xamples of the ablation study. \textit{Ours} generates robust results to different styles and draws the hair and mustache more accurately compared to the alternatives. Without $L_1$ initialization and with $L_1$ until the end, incomplete lines or artifacts on the face are generated. Without attention, the mustache and wrinkles disappear. With one adversarial loss, the hair area is not drawn the same as the ground truth. Without clip, noticeable artifacts on the face and hair appear. Utilizing only partial features cannot properly generate the first style.}
\end{figure*}

\subsection{Ablation study} \label{subsec:ablation}
To conduct an ablation study, we trained our network with different loss conditions, training scheme\fixw{s,} and architecture designs. Specifically, we trained our network without L1 loss initialization, with L1 loss until the end, without attention layers in the generator, with one adversarial loss, and without clip loss and utilizing only the partial features from StlyeGAN. \fixr{Because we did not compare the results with those from APdrawing++ for ablation, we did not align the images to the landmark control points \fixw{that were} pre-defined by APdrawing++, unlike \fixw{the} comparisons performed with baselines.} See Table~\ref{Tab:ablation2} \news{for the resulting scores which are averaged over all nine styles. \newa{Table 3 in the supplementary material} provides a comprehensive presentation of individual quantitative results for all nine styles.}

\begin{table}[t]
\centering
\small 
\setlength{\tabcolsep}{8pt} 
\renewcommand{\arraystretch}{1.2} 
\caption{Ablation study using LPIPS and FID scores, average of nine styles.} \vspace{-10pt}
\begin{tabular}{|c|cc|}
\noalign{\smallskip}\noalign{\smallskip}\hline

    Methods      &\textbf{LPIPS↓} & \underline{FID}↓ \\ \hline
    
\textit{Ours} & \textbf{0.1772} & \underline{81.39} \\
\textit{Ours} W/O $L_1$ initialization & 0.1837 & 94.98 \\
\textit{Ours} W $L_1$ until the end  & 0.1885 & 113.45 \\
\textit{Ours} W/O attention  & 0.1805 & 92.80 \\
\textit{Ours} W one adversarial loss & 0.1861 & 94.06 \\
\textit{Ours} W/O $L_{clip}$  & 0.1855 & 96.01 \\ 

\news{\textit{Ours} deep fusion module W/O $f_{i-1}$} &\news{0.2064} & \news{120.15} \\ 

\textit{Ours} W first half features &  0.2452 & 192.92 \\ 

\textit{Ours} W last half features & 0.1974 & 98.40 \\ 

\textit{Ours} W middle ten features & 0.1929 & 102.40 \\ 

\hline 
\end{tabular}
\label{Tab:ablation2}
\end{table}

In the ablation study that focused on initialization, \textit{Ours} clearly outperformed \textit{Ours} without L1 initialization and \textit{Ours} with L1 until the end, which is also verified visually in Fig.~\ref{fig:curriculum}. \textit{Ours} without L1 initialization required a significantly larger number of iterations to generate images in the sketch domain compared to \textit{Ours} with L1 until the end or \textit{Ours}. While \textit{Ours} and \textit{Ours} with L1 until the end were able to produce images that can be considered in the sketch domain within approximately 2000 iterations, \textit{Ours} without L1 initialization failed to do so even after 4000 iterations. This difficulty in optimizing the output into the sketch domain leads to poor image quality in the final result. 

\textit{Ours} without attention and \textit{Ours} with one adversarial loss resulted in cut lines or deleted information in the hair and mustache regions. Additionally, when \textit{Ours} used only the first half of the 18 available features, it generated results with missing information. \news{\textit{Ours} that used a deep fusion module without $f_{i-1}$, last features, and  middle ten features generated reasonable sketches, albeit with degraded quality, especially in the first style.} Overall, \textit{Ours} outperformed other alternatives in terms of the measured scores, which is reflected in the visual results, as shown in Fig.~\ref{fig:ablationexample}. The results produced by \textit{Ours} show superior quality in terms of \fixw{the completeness of the generated lines or faithful reconstruction of local features}, while the \fixw{results} of \fixw{the alternatives} suffer from missing lines and noise.

\begin{figure}[!t]
  \centering
  \includegraphics[width=0.9\linewidth]{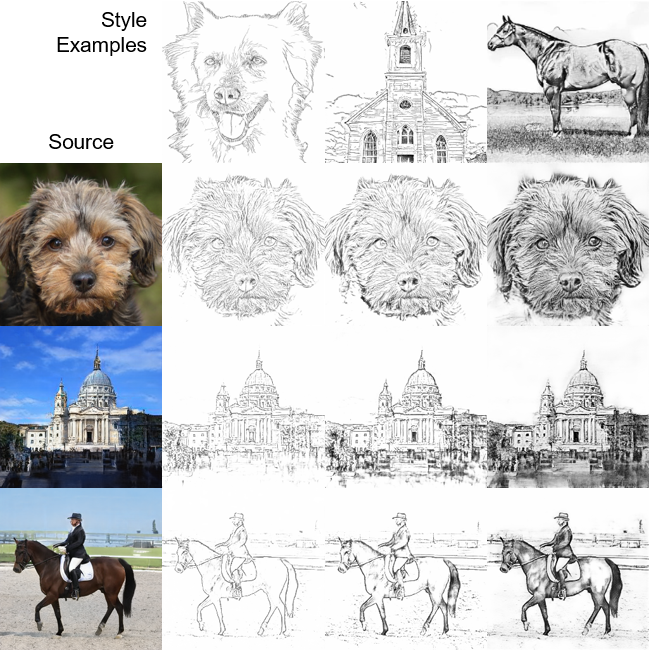}

  \caption{Results of StyleSketch in different domains. For each style, $G_{sketch}$ was trained with 16 data pairs of images generated from StyleGAN and the corresponding sketches extracted using a previous method \cite{seo2023semi}.}
  \label{fig:app_other}
  \vspace{-0.5cm}
\end{figure}

\begin{figure*}[!htb]
  \centering
  \includegraphics[width=0.9\linewidth]{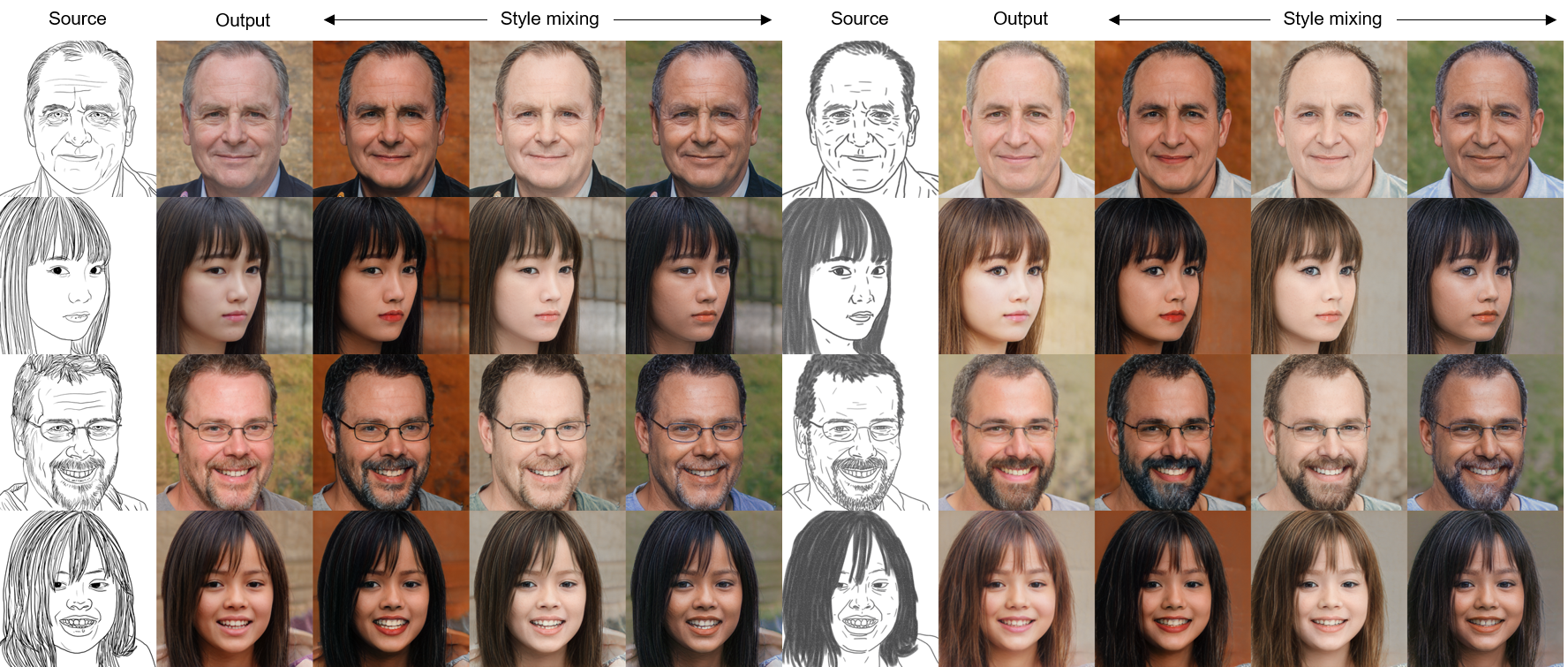}

  \caption{Synthesizing multiple photo-realistic faces from sketch images by training pSp~\cite{richardson2021encoding} with StyleSketch synthetic images.}
  \label{fig:app_sketch2face}
\end{figure*}

\section{Application}

\subsection{Sketch generation in different domain}\label{sec:other}
Although our explanation has focused on the extraction of sketches from face images, StyleSketch can easily extend to handle images from various other domains such as dogs or outdoor scenes as long as related deep features can be obtained from a pretrained StyleGAN. We tested StyleSketch with StyleGAN pretraiend with AFHQ-dog~\cite{choi2020stargan}, LSUN-horse, and LSUN-church \cite{yu2015lsun}.
We first modified the loss functions for both $\mathcal{L}_{adv}$ and $\mathcal{L}_{perc}$. While methods to parse facial parts are available, there are not many methods for segmenting parts for churches or animal faces; thus, we modified $\mathcal{L}_{adv}$ to use three discriminators, each of which is specialized for the background, foreground, and full sketch. In addition, we did not use the part-based $\mathcal{L}_{perc}$ and instead used the full image to compute $\mathcal{L}_{perc}$. With these modifications, we were able to train the sketch generator with the same number of limited data pairs which consequently enables us to perform sketch extraction, as shown in Fig.~\ref{fig:app_other}. Sketches in each column reflect the style of the corresponding training data.

\subsection{Sketch to face image generation}
\label{subsec:sketch2face}

\fixe{Photo-realistic images can be generated from a face sketch by inverting the} given sketch to the latent space of pretrained StyleGAN.
Finding the correct latent code for a given sketch is a non-trivial task. We first tested an optimization-based method~\cite{abdal2020image2stylegan++} \fixe{for the purpose}. While the optimized code was able to generate a sketch similar to the given sketch, the latent code did not lie inside the valid space of pretrained StyleGAN.
Thus, we opted to train an encoder that inverts a sketch image to a latent code. For this, we first generated a huge number of paired face and sketch image data with StyleGAN $G$ and the sketch generator $G_{sketch}$. Following the training strategy of pSp~\cite{richardson2021encoding}, the trained encoder with the pretrained StyleGAN can successfully generate photo-realistic face images given an input sketch, as shown in Fig.~\ref{fig:app_sketch2face}.
Because our method can generate pairs of face images and sketches in different styles, the trained encoder can invert the sketch, matching the identity and characteristics of both the sketch and the original source image. 

We emphasize that the ability of the encoder to successfully convert diverse style sketches to photo-realistic images is due to the use of images generated by StyleSketch. Furthermore, because the sketch does not convey the color information, it can be multi-modal. Therefore, the predicted \fixr{latent code} can be mixed with a desired style code to generate photo-realistic face images of various styles.

\begin{figure*}[h]
  \centering
  \includegraphics[width=0.9\linewidth]{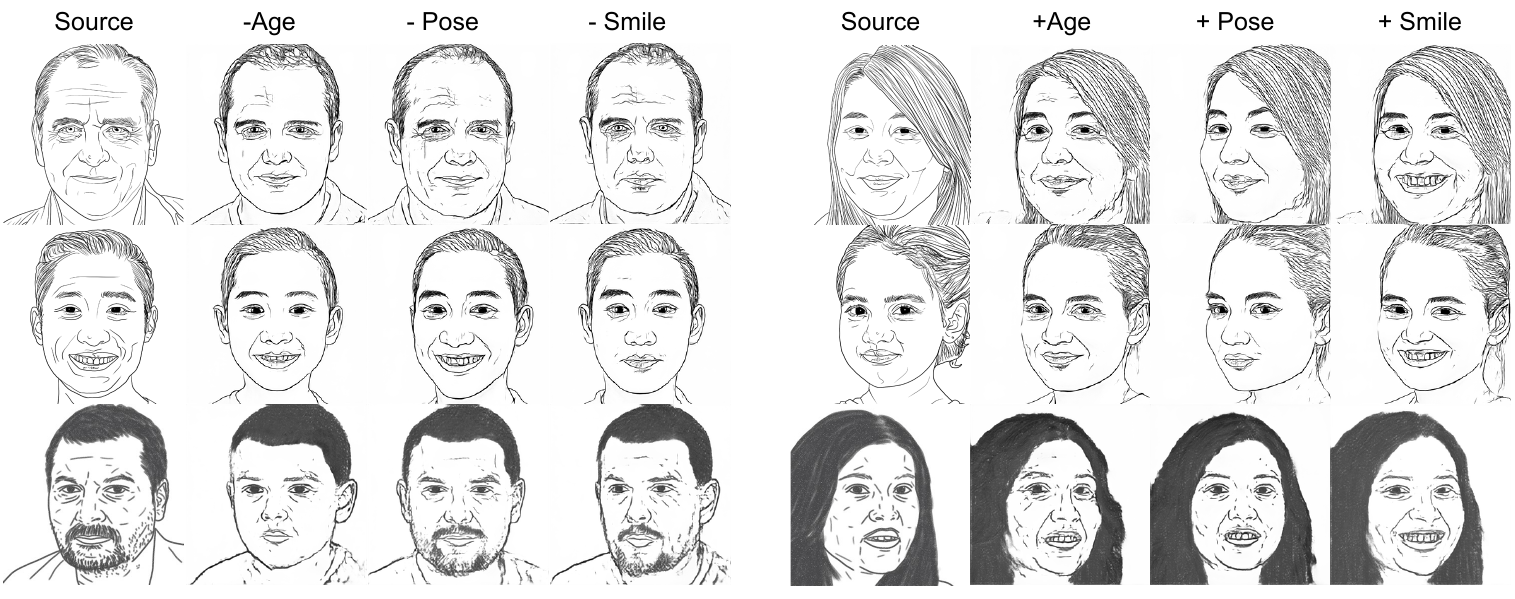}

  \caption{Semantic editing of face sketches. \newr{Given a sketch as input, semantic editing can be applied.}}
  \label{fig:app_edit}
  \vspace{-0.1cm}
\end{figure*}

\subsection{Semantic editing}
\label{subsec:editing}

\fixe{As in image manipulation tasks, users may want to edit their sketches to achieve semantically different results, such as increasing the degree of smiling or facing a different direction.} While previous methods focus only on extracting the sketch from an image, StyleSketch can easily extend to perform semantic editing using the property of the pretrained latent space. To achieve this, we extract \fixr{the latent code from the sketch image, as explained in Sec.~\ref{subsec:sketch2face}.} The extracted latent code is manipulated by adding a known semantic latent direction~\cite{shen2020interpreting, harkonen2020ganspace} followed by generating the edited image using $G_{sketch}$. Examples of edited sketches are shown in Fig.~\ref{fig:app_edit}. We observed that while the identity and style of the sketch are preserved, semantics such as age, pose, and expression can be effectively manipulated.

\section{Limitations and Future Work}

StyleSketch is capable of extracting high-resolution sketches from an input face image after training with a limited number of data samples. \news{Because StyleSketch are trained with few data, it occasionally produces blurry sketches or noisy results when discrepancy between the training and test images exists. In addition, because} StyleSketch utilizes deep features from a pretrained StyleGAN, the quality of inversion is crucial for effective training. If the inversion is not performed correctly, $G_{sketch}$ can start learning incorrectly from the beginning. \newe{This can also pose a problem at inference, because the same inversion is performed during training and inference.} This challenge originates from the GAN model and its inversion techniques. Because a distortion-perception trade-off exists when inverting an image to \news{$w^+$} latent space~\cite{tov2021designing}, \newr{a method that avoids this trade-off can reduce the dependency on the quality of inversion. For example, estimating the deep features directly during inversion, as proposed in the Feature-Style Encoder \cite{yao2022feature}, and fusing them with the generated features could be a possible solution. We leave this as an intriguing direction for future research.}

\newe{Another possible direction is to utilize distillation by training a new network to skip the inversion. Using paired sketch and face images generated from StyleSketch, an image-to-image translation network can be trained with these synthetic data \cite{viazovetskyi2020stylegan2}. Through distillation, the newly trained network will not be influenced by inversion as it directly utilizes the input image.}
\section{Conclusion}
In this study, we present a novel approach for extracting high-resolution stylized sketches from a source face image after training the network with a limited number of paired data. 
Unlike previous approaches, which directly extract sketches from images, thus providing limited information that makes training deep neural networks with a small number of paired data difficult, our method boosts the information from the inversion and deep feature extraction process. In addition, StyleSketch incorporates spatial and channel features through the implementation of a newly proposed deep fusion module. We also introduced a two-stage learning technique that is specifically designed to enable the model to be trained for sketch extraction.

We validated the effectiveness of our method through a comprehensive set of both quantitative and qualitative evaluations. \fixe{Although our explanations focused on sketch extractions in the face domain, we also showed that our method can be extended into other domains such as dog and building images using limited data. Leveraging our method, sketch-based applications such as face image generation from sketches and semantic editing is also possible without a large dataset. We believe that our method can be extensively utilized across various industries, through diverse applications, and can stimulate further research.}

\section*{Acknowledgement}
This research was supported by Culture, Sports and Tourism R\&D Program through the Korea Creative Content Agency grant funded by the Ministry of Culture, Sports and Tourism in 2023 (Project Name: Development of Universal Fashion Creation Platform Technology for Avatar Personality Expression, Project Number: RS-2023-00228331, Contribution Rate: 100\%).

\bibliographystyle{eg-alpha-doi} 
{\newcommand{\etalchar}[1]{$^{#1}$}

}

\end{document}